\pdfoutput=1

\documentclass[11pt]{article}

\usepackage[final]{acl}

\usepackage{times}
\usepackage{latexsym}

\usepackage[T1]{fontenc}

\usepackage[utf8]{inputenc}

\usepackage{microtype}

\usepackage{inconsolata}

\usepackage{graphicx}

\usepackage{inconsolata}
\usepackage{enumerate}
\usepackage{enumitem}
\usepackage{tikz}
\usepackage{amsmath}
\usepackage{subcaption}
\usepackage{amssymb}
\usepackage{booktabs}
\usepackage{arydshln}
\usepackage{xcolor}
\definecolor{customorange}{RGB}{237,125,49}
\definecolor{customblue}{RGB}{68,114,196}
\definecolor{customgreen}{RGB}{0,176,80}
\usepackage[normalem]{ulem}
\useunder{\uline}{\ul}{}
\definecolor{royalblue}{rgb}{0.25, 0.41, 0.88}

\usepackage{ulem}
\title{Coverage-based Fairness in Multi-document Summarization}

\author{
  \textbf{Haoyuan Li\textsuperscript{1}},
  \textbf{Yusen Zhang\textsuperscript{2}},
  \textbf{Rui Zhang\textsuperscript{2}},
  \textbf{Snigdha Chaturvedi\textsuperscript{1}}
\\
  \textsuperscript{1}University of North Carolina at Chapel Hill,
  \textsuperscript{2}Pennsylvania State University 
\\
  {\{haoyuanl, snigdha\}@cs.unc.edu, \{yfz5488, rmz5227\}@psu.edu}
}

\begin{document}
\maketitle
\begin{abstract}
Fairness in multi-document summarization (MDS) measures whether a system can generate a summary fairly representing information from documents with different social attribute values. Fairness in MDS is crucial since a fair summary can offer readers a comprehensive view. 
Previous works focus on quantifying summary-level fairness using Proportional Representation, a fairness measure based on Statistical Parity. 
However, Proportional Representation does not consider redundancy in input documents and overlooks corpus-level unfairness. 
In this work, we propose a new summary-level fairness measure, \textbf{Equal Coverage}, which is based on coverage of documents with different social attribute values and considers the redundancy within documents. To detect the corpus-level unfairness, we propose a new corpus-level measure, \textbf{Coverage Parity}. 
Our human evaluations show that our measures align more with our definition of fairness. Using our measures, we evaluate the fairness of thirteen different LLMs. We find that Claude3-sonnet is the fairest among all evaluated LLMs. We also find that almost all LLMs overrepresent different social attribute values. The code is available at \href{https://github.com/leehaoyuan/coverage_fairness}{https://github.com/leehaoyuan/coverage\_fairness}.
\end{abstract}

\section{Introduction}
\label{intro}
Multi-document summarization (MDS) systems summarize the salient information from multiple documents about an entity, such as news articles about an event or reviews of a product. 
Typically, such documents are associated with \textit{social attributes} e.g. political ideology in news and sentiments in reviews. 
Documents with different social attributes tend to have diverse information or conflicting opinions. 
A summary for them should fairly represent differing opinions across documents.   

Fairness in MDS measures whether a system can generate summaries fairly representing information from documents with different social attribute values. It can be measured at a \textit{summary-level}-- quantifying how fairly an individual summary represents documents with different social attribute values or at a \textit{corpus-level}--quantifying how fairly a corpus of summaries as a whole represents documents with different social attribute values.  
Previous works in this area measured fairness in extractive or abstractive settings \cite{shandilya2018fairness, olabisi-etal-2022-analyzing, zhang2023fair, huang2024bias}. These works generally evaluate the fairness of a system as the aggregated summary-level fairness of its generated summaries. 
It is measured using Proportional Representation--a fairness measure based on Statistical Parity \cite{verma2018fairness}. 
It requires that a document sentence being selected as a summary sentence is independent of its originating document's social attribute value. Therefore, the distribution of social attributes among information in a fair summary should be the same as the distribution of social attributes among input documents. 

 \begin{figure*}[t!]
     \centering
     \begin{subfigure}[t]{0.48\textwidth}
         \centering
        \includegraphics[width=\textwidth]{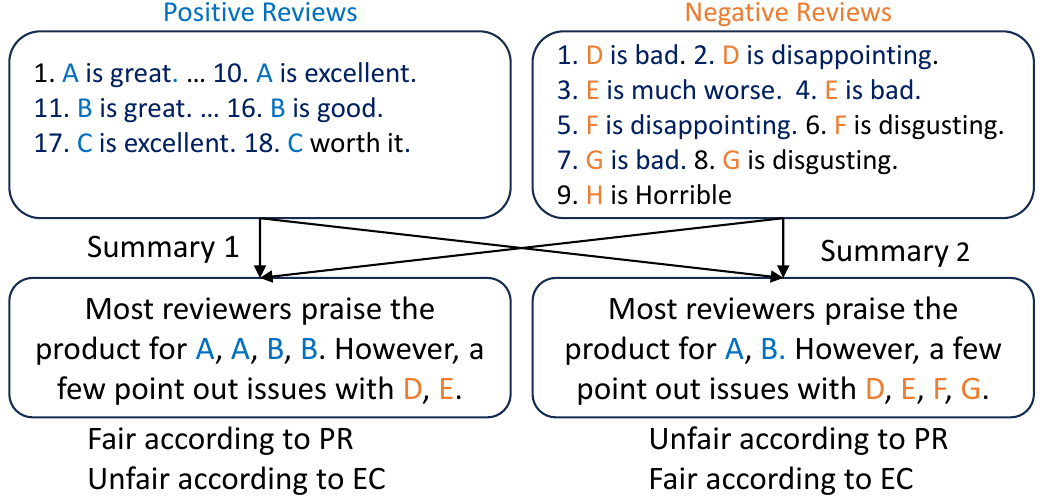}
        \caption{Existing summary-level fairness measure, Proportional Representation (PR), does not consider redundancy common in MDS. According to PR, Summary 1 is fairer than Summary 2 while Summary 2 is clearly better since it avoids repetition and cover equal proportions of information from both sentiments. Our proposed measure, Equal Coverage, correctly considers Summary 2 as fairer than Summary 1.}
         \label{fig:duplication}
     \end{subfigure}
     \hfill
     \begin{subfigure}[t]{0.48\textwidth}
         \centering
         \includegraphics[width=\textwidth]{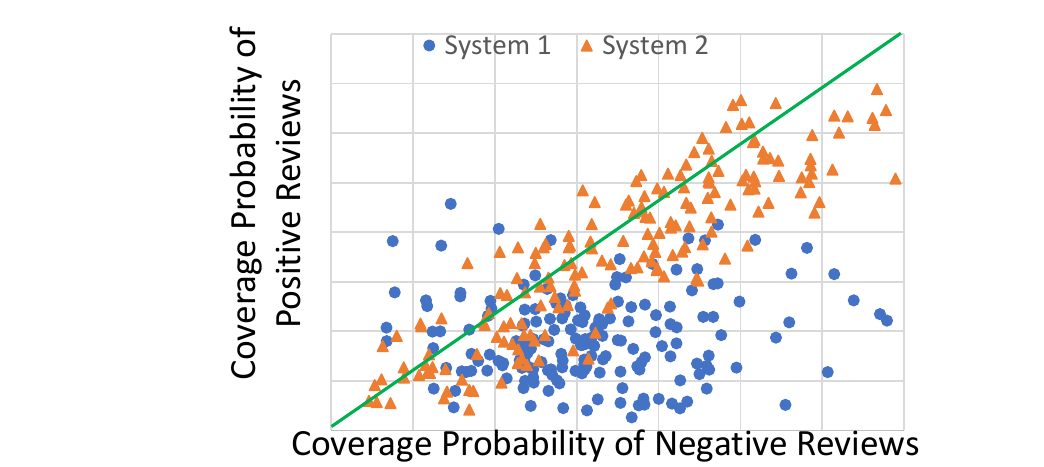}
         \caption{Existing summary-level fairness measures can overlook corpus-level unfairness. Each point in this figure represents a summary sample. \textcolor{customorange}{System 2} is fairer than \textcolor{customblue}{System 1} since it has equal chances of overrepresenting negative (below the \textcolor{customgreen}{green} line) and positive (above the \textcolor{customgreen}{green} line) reviews while System 1 tends to overrepresent negative reviews. Our proposed measure, Coverage Parity, can correctly identify \textcolor{customorange}{System 2} as fairer than \textcolor{customblue}{System 1}. }
         \label{fig:system_level}
     \end{subfigure}
        \caption{Issues with existing fairness measures for summary-level (a) and corpus-level (b) fairness.}
        \label{fig:problem}
\end{figure*}
The definition of Proportional Representation (PR) suffers from two key problems. The first problem is that Proportional Representation does not consider the redundancy in input documents, common in MDS. For example, in reviews of a hotel with great views, multiple reviews could comment on the view. For input documents with much redundancy, it is difficult to generate a non-redundant, fair summary according to PR.
For example, suppose $67\%$ of input reviews are positive and mostly discuss topics A and B, while $33\%$ of reviews are negative and evenly discuss topics D, E, F, and G (Fig. \ref{fig:duplication}).
According to PR, Summary 1 is fairer than Summary 2. However, Summary 2 is clearly better since, unlike Summary 1, it avoids repetition (Summary 1 repeats topic A twice) and almost equally \textit{cover} the information from both positive and negative reviews. Hence, there is a need to redefine fairness to consider redundancy in MDS. Besides, the idea behind PR can work for measuring fairness in other NLP tasks but not for abstractive summarization because recent LLM-based summarization methods can use quantifiers like "most" and "a few" to indicate the amount of input information \textit{covered} by a summary sentence. Hence, there is a need to define fairness based on the proportion of input information \textit{covered} by the summary. With this motivation, we propose a new summary-level coverage-based fairness measure, \textbf{Equal Coverage} (EC). Unlike PR, EC requires a document being \textit{covered} by a summary sentence to be independent of its social attribute value. Since a summary sentence can cover multiple documents with similar contents, EC can address redundancy common in MDS. 

The second problem is evaluating the fairness of an MDS system only using summary-level fairness measures can overlook corpus-level unfairness. Consider System 1 in Fig. \ref{fig:system_level}. Most of its summaries (\textcolor{customblue}{blue} dots) have a higher coverage probability for negative reviews than positive reviews. We observe that System 1 is unfair because its summaries  \textit{overrepresent} negative reviews. System 2 is fairer because its summaries (\textcolor{customorange}{orange} dots) have an equal chance of overrepresenting negative or positive reviews. Since individual summaries from both systems overrepresent negative or positive reviews, their summary-level fairness scores may be comparable. Hence, aggregated summary-level fairness scores cannot identify that System 2 is collectively fairer than System 1. To address this problem, we propose a new corpus-level fairness metric, \textbf{Coverage Parity}. Coverage Parity is based on the principle that documents with different social attribute values should have equal chances of being overrepresented or underrepresented. Therefore, it can check whether the systems are equally (un)fair on different social attributes and identify which social attribute is overrepresented or underrepresented.

Our human evaluation shows that our measures align more with our definition of fairness than Proportional Representation. Using our measures, we evaluate the fairness of thirteen different Large Language Models (LLMs) \cite{ouyang2022training, touvron2023llama} in diverse domains: news, tweets, and reviews. For these domains, we consider social attributes with significant real-world impacts:  ideologies and stances for news and tweets and sentiment for reviews. Our experiments find that for Llama2, Claude3, Mixtral, and Gemma2, larger models are fairer, but this trend is inconsistent for GPTs and Llama3.1. Our experiments also find that most LLMs tend to overrepresent certain social attribute values in each domain. It is an important finding that users can use to calibrate their perception before using LLM-generated summaries. It can also be used by developers to build fairer LLM-based summarization systems.

To conclude, our contributions are three-fold:
\begin{itemize}[topsep=1pt, leftmargin=*, noitemsep]
    \itemsep0mm
    \item We propose a new summary-level fairness measure, Equal Coverage, which incorporates redundancy of input information, common in MDS; 
    \item We propose a new corpus-level fairness measure, Coverage Parity to detect corpus-level unfairness;  
    \item We evaluate the fairness of LLMs using these two measures in various domains. 
\end{itemize}
\section{Related Work}

\citet{shandilya2018fairness, shandilya2020fairness, dash2019summarizing} propose to measure the summary-level fairness in MDS under the extractive setting using Proportional Representation.
They propose an in-processing method to improve the fairness of extractive summaries by adding a fairness constraints to the optimization target. Similarly, \citet{keswani2021dialect} uses Proportional Representation to measure the fairness of summaries under different distributions of social attributes in input documents and proposes a post-processing method to improve fairness. \citet{olabisi-etal-2022-analyzing} uses the same measure to measure fairness and proposes a clustering-based pre-processing method to improve fairness.

Recently, \citet{zhang2023fair, huang2024bias} extend Proportional Representation to measure fairness under the abstractive setting. To estimate the distribution of social attributes in a summary, \citet{zhang2023fair} maps the summary back to the originating documents, while \citet{huang2024bias} uses a finetuned model to estimate the social attribute of each summary sentence. \citet{lei2024polarity} measures the fairness similarly as \citet{huang2024bias} and proposes to improve the fairness of abstractive summaries using reinforcement learning with a polarity distance reward. However, these works generally measure fairness using Proportional Representation, which has the limitations discussed in the introduction. In our experiments, we use Proportional Representation as a baseline.

 
\section{Fairness Measures} 

In this section, we describe notation (Sec. \ref{notation}) and our proposed measures, Equal Coverage (Sec. \ref{sec:equal_coverage}) and Coverage Parity (Sec. \ref{sec:coverage_parity}). 
\subsection{Notation}
\label{notation}
We use $G$ to denote all samples for evaluating the fairness of a system on a social attribute. 
 Each sample $(D,S) \in G$ contains a document set $D=\{d_1,...,d_n\}$ and a summary $S$ generated by the system for these documents, where $d_i$ denotes the $i$-th document. 
 Each input document $d_i$ is labeled with an social attribute value $a_i \in \{1,...,K\}$.
\subsection{Equal Coverage}
\label{sec:equal_coverage}
Equal Coverage is a summary-level fairness measure for measuring the fairness of a summary, $S$, for document sets, $D$. 
Equal Coverage is based on the principle that the documents with different social attribute values should have equal probabilities for being covered by a summary sentence. We denote the probability that a random document $d \in D$ whose social attribute value $a$ is $k$ is covered by a random summary sentence $s \in S$ as $p(d,s|a=k)$. This is referred to as the coverage probability for the social attribute value $k$. Similarly, we denote the probability that a random document, $d$, is covered by a random summary sentence $s \in S$ as $p(d,s)$, which is referred to as the coverage probability for a document. 
For a fair summary $S$ according to Equal Coverage, coverage of a random document $d$ should be independent of its social attribute value: 
\vspace{-0.15cm}
\begin{equation}
p(d,s|a=k)=p(d,s),\ \forall k.
\end{equation}
\vspace{-0.10cm}

However, two issues arise with summaries' complex sentence structures. First, a summary sentence can combine information from several documents, making it difficult to attribute the sentence to any single document. Second, sentence lengths vary significantly based on social attribute values. 
The length difference can skew the coverage probability for different social attribute values $p(d,s|a=k)$. To address these issues, Equal Coverage splits the summary sentences $s\in S$ into multiple simpler sentences by prompting GPT-3.5 motivated by \citet{bhaskar2023prompted, min2023factscore}. For example, compound sentences are split into simple sentences that describe a single fact. To reduce redundancy among split sentences, we filter out sentences with similar meanings based on entailment between split sentences. We denote the $j$-th summary sentence after split as $s_j$. Further details are in App. \ref{sec:split_and_rephrase}.

Equal Coverage estimates the coverage probability for different social attribute values $p(d,s|a=k)$ and for a document $p(d,s)$ based on the probability $p(d_i,s_j)$ that a document $d_i$ is covered by a summary sentence $s_j$. The probability $p(d_j,s_k)$ is estimated as the entailment probability that the document $d_j$ entails the summary sentence $s_k$ by a textual entailment model \cite{N18-1101}. Motivated by \citet{laban2022summac}, Equal Coverage divides documents into chunks of $M$ words. The $l$-th chunk of the document $d_i$ is denoted as $d_{i,l}$. The entailment model estimates the probability $p(d_i,s_j)$ as the maximum entailment probability $p(d_{i,l}, s_j)$ between the chunk $d_{i,l}$ and the summary sentence $s_j$:
\vspace{-0.15cm}
\begin{equation}
p(d_i,s_j)=\max\{p(d_{i,l},s_j)|d_{i,l}\in d_i\},
\label{equ:entail}
\end{equation}
\vspace{-0.10cm}
where $p(d_{i,l},s_j)$ is the probability that the document chunk $d_{i,l}$ entails the summary sentence $s_j$ as per the entailment model. Based on the probability $p(d_i,s_j)$ that the document $d_i$ is covered by the summary sentence $s_j$, the coverage probability for social attribute value $i$, $p(d,s|a=k)$ is empirically estimated as: 
\vspace{-0.15cm}
\begin{equation}
p(d,s|a=k)=\frac{1}{|D_k||S|} \sum_{d_i \in D_k} \sum_{s_j\in S} p(d_i,s_j),
\vspace{-0.05cm}
\end{equation}
where $D_k$ is the set of documents $d_i$ whose social attribute value $a_i$ is $k$. Similarly, the coverage probability for a document, $p(d,s)$ is estimated as: 
\vspace{-0.15cm}
\begin{equation}
p(d,s)=\frac{1}{|D||S|} \sum_{d_i\in D} \sum_{s_j\in S} p(d_i,s_j).
\vspace{-0.10cm}
\label{eqn:cov_prob}
\end{equation}

Recall that for a fair summary $S$ according to Equal Coverage, coverage probabilities for different social attribute values $p(d,s|a=k)$ should equal the coverage probability for a document $p(d,s)$. Therefore, Equal Coverage measures the summary-level fairness $EC(D,S)$ as:
\vspace{-0.15cm}
\begin{equation}
EC(D,S)=\frac{1}{K} \sum_{k=1}^{K} |p(d,s)-p(d,s|a=k)|.
\end{equation} 
A high Equal Coverage value $EC(D,S)$ indicates less fairness because there are big differences between coverage probabilities for different social attribute values $p(d,s|a=k)$. 


To evaluate the fairness of the system, we use the average Equal Coverage value $EC(G)=\frac{1}{|G|}\sum_{ (D,S)\in G}EC(D,S)$ of all examples.
\subsection{Coverage Parity}
\label{sec:coverage_parity}
Coverage Parity is a corpus-level fairness measure designed to measure the fairness of a system of all samples $G$. Coverage Parity is based on the principle that documents with different social attribute values should have equal chances of being overrepresented or underrepresented by their corresponding summaries. 

For a fair MDS system, Coverage Parity requires that the average difference between the coverage probability for the document $d_i$, $p(d_i,s)$, and the coverage probability for a document, $p(d,s)$, (Eqn. \ref{eqn:cov_prob}) is close to zero for all documents $d_i$ whose social attribute value $a_i$ is $k$. The coverage probability for the document $d_i$, $p(d_i,s)$ is estimated as:

\begin{equation}
\vspace{-0.15cm}
p(d_i,s)=\frac{1}{|S|} \sum_{s_j\in S} p(d_i,s_j),
\vspace{-0.05cm}
\end{equation}
For simplicity, we denote the coverage probability difference between $p(d_i,s)$ and $p(d,s)$ as $c(d_i)$. We collect these coverage probabilities differences $c(d_i)$ from all input documents of the dataset $G$ whose social attribute value is $k$ into a set $C_k$.

Based on the average of the set of coverage probability differences $C_i$, Coverage Parity measures the fairness of the MDS system:
\begin{equation}
\vspace{-0.2cm}
CP(G)= \frac{1}{K}\sum_{k=1}^{K}|\mathbb{E}(C_k)|,
\vspace{-0.05cm}
\label{equ:cp}
\end{equation}
where $\mathbb{E}(C_k)$ denotes the average coverage probability difference of $C_k$. A high Coverage Parity value $CP(G)$ indicates less fairness since it suggests that the chances of being overrepresented or underrepresented are very different for some social attribute values. 
Based on whether the average coverage probability differences, $\mathbb{E}(C_k)$, is greater or less than zero, we can also identify which social attribute value tends to be overrepresented or underrepresented. 
\begin{table*}[t]
\centering
\small
\setlength{\tabcolsep}{1mm}
\resizebox{0.98\textwidth}{!}{
\begin{tabular}{lcccccc}
\toprule
                        & Domain & Social Attribute     & Social Attribute Values         & Inp. Doc. Set Size & Avg. Doc. Len. & Avg. Doc. Len. Dif. \\ 
\midrule
Amazon                  & Review & Sentiment            & \{negative, neutral, positive\} & 8            & 40    &6        \\
MITweet                 & Tweet  & Political Ideology   & \{left, center, right\}         & 20           & 34     & 3        \\
Article Bias  & News   & Political Ideology   & \{left, center, right\}         & 4-8          & 436    & 19       \\
SemEval                 & Tweet  & Stance toward Target & \{support, against\}            & 30           & 17     & 1       \\
News Stance    & News   & Stance toward Target  & \{support, against\}            & 4-8          & 240     & 35      \\ 
\bottomrule
\end{tabular}}
\caption{Dataset statistics for fairness evaluation in MDS. We report each dataset's domain, social attribute, size of input document set (Inp. Doc. Set Size), average length of input documents (Avg. Doc. Len.), and average length difference between input documents with different social attribute values (Avg. Doc. Len. Dif.).}
\label{tab:dataset}
\end{table*}

\section{Experimental Setup} 
We now describe experimental setup to evaluate the fairness of LLMs using our measures.
\subsection{Datasets}
\label{sec:dataset_main}
We conduct experiments on five different datasets from the three domains: reviews, tweets, and news. 
These datasets are the Amazon \cite{ni-etal-2019-justifying}, MITweet \cite{liu2023ideology}, SemEval \cite{mohammad-etal-2016-semeval}, Article Bias \cite{baly2020we}, and News Stance \cite{ferreira2016emergent, pomerleau2017fake,hanselowski2019richly} datasets. Tab. \ref{tab:dataset} shows the statistics of these datasets along with their social attribute values.

We observe that for some datasets, the fairness of summaries depends on the distributions of social attribute values in the input documents (Sec. \ref{sec:distribution}).
To balance social attribute values' impacts on fairness, we perform stratified sampling to collect $300$ input document sets, $D$, for each dataset. 
Among these sampled sets, input document sets $D$ dominated by different social attribute values have equal proportions. The stratified sampling does not affect the calculation of our fairness measures. More details of preprocessing are in App. \ref{sec:dataset}.
\subsection{Implementation Details}
We evaluate the fairness of six families of LLMs: GPTs \cite{ouyang2022training, achiam2023gpt} (GPT-3.5-0124, GPT-4-turbo-2024-04-09), Llama2 \cite{touvron2023llama} (Llama2-7b, Llama2-13b, Llama2-70b), Llama3.1 \cite{llama3modelcard} (Llama3.1-8b, Llama3.1-70b), Mixtral \cite{jiang2023mistral,jiang2024mixtral} (Mistral-7B-Instruct-v0.1, Mixtral-8x7B-Instruct-v0.1), Gemma2 \cite{team2024gemma} (gemma-2-9b-it, gemma-2-27b-it), and Claude3 \cite{bai2022constitutional} (claude-3-haiku-20240307, claude-3-sonnet-20240229).
For comparison, we also evaluate the fairness of COOP \cite{iso2021convex} on the Amazon dataset and the fairness of PEGASUS \cite{zhang2020pegasus} and PRIMERA \cite{xiao-etal-2022-primera} finetuned on the Multi-News \cite{fabbri-etal-2019-multi} on the other datasets. The summary length is 100 words for the Article Bias and News Stance datasets and 50 for the other datasets. We describe summarization prompts in App. \ref{sec:sum_prompt}.

To measure the difficulty of obtaining fair summaries on different datasets, we estimate lower bounds, Lower\textsubscript{gre}, and upper bounds, Upper\textsubscript{gre}, for both fairness measures by greedily extracting sentences that minimize or maximize the measures. We also estimate random bounds, Random, by randomly extracting sentences from input document sets as summaries. Please note that Lower\textsubscript{gre} and Upper\textsubscript{gre} are not baselines but serve as empirical bounds for our measures. The details of these bounds are in App. \ref{sec:bound}.


To estimate the probability that a document is covered by a summary sentence (Eqn. \ref{equ:entail}), we use RoBERTa-large finetuned on the MNLI dataset \cite{N18-1101}. However, our measures are independent of this choice. We show the correlation between measures calculated using other entailment models \cite{schuster-etal-2021-get,laurer2024less} in App. \ref{sec:entail}. To estimate the entailment probability using RoBERTa-large, we divide documents into chunks of $W=100$ words. The chunk size $W$ is tuned to maximize the proportion of summary sentences whose originating documents can be identified by RoBERTa-large (App. \ref{sec:chunk}) since LLMs are less prone to factual errors \cite{goyal2022news}.

\section{Experiment Results}
We now describe experiment results to evaluate the fairness of LLMs using our measures.
\subsection{Human Evaluation}
\label{sec:human_eval}
We perform a human evaluation to determine which measure, Equal Coverage or Proportional Representation, aligns more with our definition of fairness, which is to fairly represent information from documents with different social attribute values. For Proportional Representation, we use the BARTScore \cite{NEURIPS2021_e4d2b6e6} implementation proposed by \citet{zhang2023fair}, as it shows the highest correlation with human perception.

Performing human evaluation for fairness on summarization is challenging due to the need to read entire input document sets. Therefore, we perform experiments on the Amazon dataset which only contains eight short reviews per input document set (Tab. \ref{tab:dataset}). Besides, it is easier for people without training to judge whether an opinion is positive or negative than to judge if it is left-leaning or right-leaning. To simplify the evaluation, we only consider input document sets with only negative or positive reviews, displayed in two randomized columns. For each input document set, we consider the summary generated by GPT-3.5 since it shows medium-level fairness (Tab. \ref{tab:equal_coverage}, \ref{tab:cp}). To further simplify the evaluation, we focus on summaries where Equal Coverage and Proportional Representation disagree on their fairness. We randomly select $25$ samples containing input document sets and corresponding summaries that meet these criteria. Each sample is annotated by three annotators recruited from Amazon Mechanical Turk. The annotators should be from English-speaking countries and have HIT Approval Rates greater than $98\%$. More details are in App. \ref{app:human}.

Following previous practice of performing human evaluation on fairness in summarization \cite{shandilya2020fairness}, annotators are asked to read all reviews and identify all unique opinions. They then are asked to read the summary and rate its fairness as leaning negative, fair, or leaning positive. 
The Randolph’s Kappa \cite{randolph2005free} between annotations of three annotators is $0.42$, which shows moderate correlation. Out of these $25$ samples, annotations aligns more with Equal Coverage in $17$ samples, while aligns more with Proportional Representation in $8$ samples. The difference is statistically significant ($p<0.05$) using paired bootstrap resampling \cite{koehn-2004-statistical}. It shows that our implementation aligns more with our definition of fairness than Proportional Representation. 

We compare Coverage Parity with second-order fairness \cite{zhang2023fair} based on Proportional Representation to evaluate their effectiveness in identifying overrepresented sentiments on the group level. For this, we perform bootstrapping to generate $5000$ groups of bootstrap samples. We find that Coverage Parity aligns with human annotation in $95\%$ of the groups, while second-order fairness aligns in $5\%$ of the groups. It shows that Coverage Parity aligns more with our definition of fairness than the second-order fairness.

\subsection{Summary-level Fairness Evaluation} 

\begin{table}[t]
\centering
\small
\setlength{\tabcolsep}{1mm}
\resizebox{0.48\textwidth}{!}{
\begin{tabular}{lcccccc}
\toprule
             & Ama.         & MIT.        & Art. Bia.   & Sem.        & New. Sta.    & Ove.        \\ \midrule
Lower\textsubscript{gre} & 0.017 & 0.016 & 0.015 & 0.003 & 0.010 & - \\
Upper\textsubscript{gre} & 0.251 & 0.133 & 0.302 & 0.072 & 0.358 & - 
\\ 
Random & 0.062 & 0.028 & 0.078 & 0.013 & 0.018 & - \\
\hdashline
GPT-3.5        & 0.079          & 0.044          & \textbf{0.078} & 0.032          & 0.129          & 0.446          \\
GPT-4          & 0.079          & 0.047          & 0.088          & 0.029          & 0.118          & 0.444          \\ \hdashline
Llama2-7b      & 0.088          & 0.046          & 0.098          & 0.032          & 0.132          & 0.712          \\
Llama2-13b     & 0.074          & 0.047          & 0.094          & 0.033          & 0.116          & 0.501          \\ 
Llama2-70b     & 0.074          & 0.048          & 0.108          & 0.030          & 0.118          & 0.520          \\ \hdashline
Llama3.1-8b    & 0.081          & 0.043          & 0.094          & 0.026          & 0.123          & 0.304          \\
Llama3.1-70b   & 0.077          & 0.044          & 0.095          & 0.030          & 0.115          & 0.341          \\ \hdashline
Mistral-7b     & 0.083          & 0.048          & 0.111          & 0.032          & 0.130          & 0.773          \\
Mixtral-8x7b   & \textbf{0.068} & 0.043          & 0.127          & 0.029          & 0.119          & 0.356          \\ \hdashline
Gemma2-9b      & 0.084          & \textbf{0.043} & 0.091          & 0.026          & \textbf{0.112} & 0.212          \\
Gemma2-27b     & 0.075          & 0.043          & 0.091          & 0.026          & 0.116          & \textbf{0.182} \\ \hdashline
Claude3-haiku  & 0.079          & 0.047          & 0.085          & 0.034          & 0.142          & 0.709          \\
Claude3-sonnet & 0.077          & 0.043          & 0.087          & \textbf{0.025} & 0.128          & 0.251         \\ \hdashline
COOP         & 0.146  & -       & -            & -       & -           & -       \\
PEGASUS      & -      & 0.052   & 0.126        & 0.027   & 0.134       & -       \\
PRIMERA      & -      & 0.056   & 0.131        & 0.025   & 0.137       & - \\     
\bottomrule
\end{tabular}}
\caption{Average summary-level fairness and overall scores on different datasets according to Equal Coverage. A lower value indicates a fairer system. \textbf{Bold} indicates the fairest system.}
\label{tab:equal_coverage}
\end{table}

To evaluate the summary-level fairness of different LLMs, we report the average Equal Coverage values for all samples in each dataset, $EC(G)$. We also report an \textit{Overall} score which is the average of normalized $EC(G)$ ($[0,1]$) using min-max normalization on all datasets. Results are in Tab. \ref{tab:equal_coverage}. 

From the table, we observe that Gemma2-27b is the fairest based on the Overall score. Among smaller LLMs (with 7 to 9 billions parameters), Gemma2-9b is the fairest.
We also observe that almost all evaluated LLMs are fairer than COOP on the Amazon dataset, and PEGASUS and PRIMERA on the MITweet and Article Bias datasets. For the comparison within families of GPTs, Llama2, Claude3, Mixtral, and Gemma2, we observe that larger models are generally fairer.

Previous works by \citet{zhang2023fair} and \citet{lei2024polarity} evaluate the fairness of LLMs using Proportional Representation. When evaluating the fairness on sentiments in the review domain, our results using Equal Coverage are consistent with these works in the finding that GPT-4 is fairer than GPT-3.5, and both are fairer than COOP. However, they find that Llama2-13b is the fairest, while our results show that Llama2-13b and Llama2-70b are comparably fair. 
When evaluating the fairness on political ideologies in the tweet domain, our results are consistent with these works on that smaller models are fairer for Llama2. However, they find that GPT-4 is fairer than GPT-3.5 while we find that GPT-3.5 is fairer than GPT-4. We also show that GPT-4 is fairer than Llama2-7b, which contrasts the previous works. We additionally report the results for Proportional Representation on our datasets in App. \ref{sec:comp_pr} and also find much inconsistency between Equal Coverage and Proportional Representation. As shown in Sec. \ref{intro}, Equal Coverage considers redundancy common in MDS, suggesting better reflects fairness in MDS.

\subsection{Corpus-level Fairness Evaluation}
\begin{table*}[h]
\small
\setlength{\tabcolsep}{1mm}
\resizebox{0.98\textwidth}{!}{
\begin{tabular}{lcccccccccccccccc}
\toprule
             & \multicolumn{3}{c}{Ama.}                       & \multicolumn{3}{c}{MIT.}                & \multicolumn{3}{c}{Art. Bia.}          & \multicolumn{3}{c}{Sem.}                    & \multicolumn{3}{c}{New. Sta.}                & \multicolumn{1}{l}{Ove.} \\
             & $CP(G)$        & over           & under          & $CP(G)$        & over       & under        & $CP(G)$        & over        & under      & $CP(G)$        & over          & under         & $CP(G)$        & over          & under         &                             \\ \midrule
Lower\textsubscript{gre}         & 0.000          & neu.       & pos.       & 0.000          & lef        & cen.       & 0.000             & rig.         & lef.       & 0.000          & aga.       & sup.       & 0.000            & sup.         & aga.       & -              \\
Upper\textsubscript{gre}         & 0.141          & {\ul neg.} & {\ul pos.} & 0.048          & {\ul rig.} & {\ul lef.} & 0.195             & {\ul rig.}   & {\ul lef.} & 0.040          & {\ul sup.} & {\ul aga.} & 0.256            & {\ul aga.}   & {\ul sup.} & -              \\ 
Random & 0.008 & neg. & neu. & 0.003 & rig. &cen. & 0.005& rig.& cen. & 0.001 &  sup. & aga. & 0.001 & sup. & aga. & - \\ \hdashline
GPT-3.5        & 0.019          & {\ul neg.} & {\ul pos.} & 0.006          & {\ul lef}  & {\ul cen.} & 0.011             & cen.         & {\ul lef}  & 0.009          & {\ul sup.} & {\ul aga.} & 0.016            & {\ul aga.}   & {\ul sup.} & 0.598          \\
GPT-4          & 0.020          & {\ul neg.} & {\ul pos.} & 0.004          & {\ul lef}  & {\ul cen.} & 0.015             & {\ul rig.}   & {\ul lef.} & 0.008          & {\ul sup.} & {\ul aga.} & 0.033            & {\ul aga.}   & {\ul sup.} & 0.622          \\ \hdashline
Llama2-7b      & 0.022          & {\ul neg.} & {\ul pos.} & 0.006          & {\ul lef}  & {\ul cen.} & 0.017             & {\ul rig.}   & {\ul lef.} & 0.008          & {\ul sup.} & {\ul aga.} & 0.011            & aga.         & sup.       & 0.634          \\
Llama2-13b     & 0.010          & {\ul neg.} & {\ul pos.} & 0.004          & {\ul lef}  & cen.       & 0.015             & {\ul rig.}   & {\ul lef.} & 0.011          & {\ul sup.} & {\ul aga.} & 0.005            & aga.         & sup.       & 0.399          \\
Llama2-70b     & \textbf{0.008} & {\ul neg.} & {\ul neu.} & \textbf{0.003} & lef        & cen.       & 0.015             & {\ul rig.}   & {\ul lef.} & 0.007          & {\ul sup.} & {\ul aga.} & \textbf{0.002}   & aga.         & sup.       & \textbf{0.182} \\ \hdashline
Llama3.1-8b    & 0.021          & {\ul neg.} & {\ul pos.} & 0.004          & {\ul lef}  & {\ul cen.} & 0.013             & {\ul rig.}   & {\ul lef.} & 0.008          & {\ul sup.} & {\ul aga.} & 0.006            & aga.         & sup.       & 0.442          \\
Llama3.1-70b   & 0.016          & {\ul neg.} & {\ul pos.} & 0.006          & {\ul lef}  & {\ul cen.} & 0.016             & {\ul rig.}   & {\ul lef.} & 0.004          & {\ul sup.} & {\ul aga.} & 0.004            & aga.         & sup.       & 0.436          \\ \hdashline
Mistral-7b     & 0.016          & {\ul neg.} & {\ul pos.} & 0.005          & {\ul lef}  & {\ul cen.} & 0.017             & {\ul rig.}   & {\ul lef.} & 0.010          & {\ul sup.} & {\ul aga.} & 0.015            & {\ul aga.}   & {\ul sup.} & 0.574          \\
Mixtral-8x7b   & 0.011          & {\ul neg.} & neu.       & 0.006          & {\ul lef}  & {\ul cen.} & 0.023             & {\ul rig.}   & {\ul lef.} & 0.002          & sup.       & aga.       & 0.020            & {\ul aga.}   & {\ul sup.} & 0.494          \\ \hdashline
Gemma2-9b      & 0.022          & {\ul neg.} & {\ul pos.} & 0.006          & {\ul lef}  & {\ul cen.} & 0.014             & {\ul cen.}   & {\ul lef}  & 0.008          & {\ul sup.} & {\ul aga.} & 0.012            & {\ul aga.}   & {\ul sup.} & 0.635          \\
Gemma2-27b     & 0.021          & {\ul neg.} & {\ul pos.} & 0.007          & {\ul lef}  & {\ul cen.} & \textbf{0.010}    & {\ul rig.}   & {\ul lef.} & 0.006          & {\ul sup.} & {\ul aga.} & 0.010            & aga.         & sup.       & 0.517          \\ \hdashline
Claude3-hai.  & 0.016          & {\ul neg.} & {\ul pos.} & 0.004          & {\ul lef}  & {\ul cen.} & 0.010             & {\ul cen.}   & {\ul lef}  & 0.010          & {\ul sup.} & {\ul aga.} & 0.037            & {\ul aga.}   & {\ul sup.} & 0.531          \\
Claude3-son. & 0.017          & {\ul neg.} & {\ul pos.} & 0.004          & lef        & {\ul cen.} & 0.012             & {\ul rig.}   & {\ul lef.} & \textbf{0.002} & sup.       & aga.       & 0.025            & {\ul aga.}   & {\ul sup.} & 0.313          \\ \hdashline
COOP           & 0.048          & {\ul pos.} & {\ul neg.} & -              & \textbf{-} & \textbf{-} & -                 & \textbf{-}   & \textbf{-} & -              & -          & -          & -                & -            & -          & -              \\
PEGASUS        & -              & \textbf{-} & \textbf{-} & 0.001          & lef        & cen.       & 0.009             & rig.         & lef.       & 0.002          & sup.       & aga.       & 0.012            & sup.         & aga.       & -              \\
PRIMERA        & -              & \textbf{-} & \textbf{-} & 0.002          & rig.       & cen.       & 0.020             & {\ul rig.}   & {\ul lef.} & 0.000          & sup.       & aga.       & 0.002            & sup.         & aga.       & -                                        \\ \bottomrule

\end{tabular}}
\caption{Coverage Parity, $CP(G)$, and the most overrepresented (over) and underrepresented (under) social attribute values, and overall scores on different datasets. A lower value of $CP(G)$ indicates a fairer system. \textbf{Bold} indicates the fairest system. The social attribute values whose average coverage probability differences are statistically significantly ($p<0.05$) different from zero based on Bootstrapping are \underline{underlined}. Most LLMs overrepresent and underrepresent certain social attribute values on different datasets}
\label{tab:cp}
\end{table*}
To evaluate the corpus-level fairness of different LLMs, we report the Coverage Parity, $CP(G)$, on different datasets. For each dataset, we report the most overrepresented and underrepresented social attribute value $k$ whose average coverage probability difference, $\mathbb{E}(C_k)$, is the maximum and minimum respectively. We also report an \textit{Overall} score which is the average of normalized $CP(G)$ ($[0,1]$) using min-max normalization on all datasets. The results are in Tab. \ref{tab:cp}.

From the table, we observe that Llama2-70b is the fairest based on the Overall score. Among smaller LLMs (with 7 to 9 billions parameters), Llama3.1-8b is the fairest. We also observe that evaluated LLMs are less fair than PEGASUS and PRIMERA on the SemEval dataset. 
While comparing within each family of LLMs, we observe that larger models are fairer for the families of Llama2, Llama3.1, Mixtral, Gemma2, and Claude3. However, for the family of GPTs, GPT-4 is less fair than GPT-3.5, suggesting that larger models are not necessarily more fair on the corpus level. Besides, we observe that the fairness measured by Coverage Parity and Equal Coverage are different on some datasets. The difference indicates that we should consider both summary-level fairness and corpus-level fairness for comprehensively measuring fairness in MDS. 

We can also observe that most LLMs overrepresent and underrepresent certain social attribute values on different datasets. For the Amazon dataset, most LLMs overrepresent negative reviews and underrepresent positive reviews. Contrarily, COOP overrepresents positive reviews and underrepresents negative reviews. For the MITweet and Article Bias datasets, all LLMs overrepresent left-leaning documents in the tweet domain, while most LLMs overrepresent right-leaning documents in the news domain. Contrarily, PEGASUS and PRIMERA overrepresent right-leaning documents for all domains. We can observe the same pattern for the SemEval and News Stance datasets. All LLMs overrepresent documents supporting the target in the tweet domain but overrepresent documents against the target in the news domain. These results indicate that summaries generated by LLMs might overrepresent documents with certain social attribute values. Users should know this before they make judgments based on these summaries. For example, users should know that a review summary generated by LLMs for a product might unfairly overrepresent negative reviews so they can calibrate their perception of the product accordingly. Developers can also build fairer LLMs for MDS based on these results.

\subsection{Fainess under Different Distributions of Social Attributes}
\label{sec:distribution}

\begin{table}[t]
\small
\setlength{\tabcolsep}{1mm}
\resizebox{0.48\textwidth}{!}{
\begin{tabular}{lccccc}
\toprule
             & Ama.      & MIT.     & Art. Bia.         & Sem.     & New. Sta. \\ \midrule
GPT-3.5      & \underline{0.023}       & 0.004       & 0.005        & 0.004       & \underline{0.022}       \\
GPT-4        & \underline{0.023}       & \underline{0.009}       & \underline{0.015}        & \underline{0.007}       & \underline{0.018}       \\ \hdashline
Llama2-7b    & \underline{0.023} & \underline{0.008}       & \underline{0.018}        & 0.004       & 0.014       \\
Llama2-13b   & \underline{0.012}       & 0.006       & 0.010        & 0.004       & 0.004       \\
Llama2-70b   & \underline{0.020} & 0.001       & 0.009  & 0.003       & 0.010       \\ \hdashline
Llama3.1-8b    & \underline{0.021} & \underline{0.005}       & 0.008        & 0.002       & \underline{0.031}       \\
Llama3.1-70b    & \underline{0.019} & \underline{0.007}       & 0.010        & 0.003       & 0.020       \\ \hdashline
Mistral-7b   & \underline{0.020}       & 0.005 & 0.018        & 0.001 & 0.012       \\
Mixtral-8x7b & \underline{0.010}       & 0.003       & 0.003        & 0.004       & 0.012       \\ \hdashline
Gemma2-9b        & \underline{0.017}       & \underline{0.007}       & 0.010        & 0.003       & \underline{0.020}       \\ 
Gemma2-27b        & \underline{0.019}       & 0.003       & 0.003        & 0.001       & 0.004       \\ \hdashline
Claude3-hai.  & \underline{0.015}       & 0.006       & 0.004       & 0.001       & 0.008       \\
Claude3-son. & \underline{0.016} & 0.003       & 0.060       & 0.004       & \underline{ 0.028} \\
\bottomrule
\end{tabular}}
\caption{Differences of Equal Coverage under different distributions of social attribute values in input document sets. Statistically significant ($p<0.05$) differences based on Bootstrapping are \underline{underlined}. We observe significant differences in summary-level fairness for most LLMs on the Amazon dataset. }
\label{tab:ec_diff}
\end{table}

\begin{table}[t]
\small
\setlength{\tabcolsep}{1mm}
\resizebox{0.48\textwidth}{!}{
\begin{tabular}{lccccc}
\toprule
             & Ama.      & MIT.     & Art. Bia.         & Sem.     & New. Sta. \\ \midrule
             
GPT-3.5        & {\ul 0.064} & {\ul 0.017} & {\ul 0.040} & {\ul 0.011} & {\ul 0.017} \\
GPT-4          & {\ul 0.074} & {\ul 0.012} & {\ul 0.062} & {\ul 0.011} & {\ul 0.012} \\ \hdashline
Llama2-7b      & {\ul 0.087} & {\ul 0.018} & {\ul 0.032} & {\ul 0.014} & {\ul 0.018} \\
Llama2-13b     & {\ul 0.064} & {\ul 0.012} & {\ul 0.040} & {\ul 0.017} & {\ul 0.012} \\
Llama2-70b     & {\ul 0.073} & {\ul 0.011} & {\ul 0.034} & {\ul 0.012} & {\ul 0.011} \\ \hdashline
Llama3.1-8b    & {\ul 0.079} & {\ul 0.012} & {\ul 0.045} & {\ul 0.014} & {\ul 0.012} \\
Llama3.1-70b   & {\ul 0.083} & {\ul 0.019} & {\ul 0.029} & {\ul 0.010} & {\ul 0.019} \\ \hdashline
Mistral-7b     & {\ul 0.062} & {\ul 0.013} & {\ul 0.046} & {\ul 0.013} & {\ul 0.013} \\
Mixtral-8x7b   & {\ul 0.056} & {\ul 0.013} & {\ul 0.036} & {\ul 0.007} & {\ul 0.013} \\ \hdashline
Gemma2-9b      & {\ul 0.085} & {\ul 0.010} & 0.033       & {\ul 0.009} & {\ul 0.010} \\
Gemma2-27b     & {\ul 0.080} & {\ul 0.021} & {\ul 0.027} & {\ul 0.013} & {\ul 0.021} \\ \hdashline
Claude3-hai.  & {\ul 0.079} & {\ul 0.013} & {\ul 0.034} & {\ul 0.011} & {\ul 0.013} \\
Claude3-son. & {\ul 0.072} & {\ul 0.010} & {\ul 0.033} & {\ul 0.012} & {\ul 0.010} \\ 
\bottomrule
\end{tabular}}
\caption{Differences of Coverage Parity under different distributions of social attribute values in input document sets. Statistically significant ($p<0.05$) differences based on Bootstrapping are \underline{underlined}. We observe significant differences in Coverage Parity for most LLMs. It indicates that the social attribute values overrepresented or underrepresented change significantly under different distributions.}
\label{tab:cp_diff}
\end{table}

\begin{table}[t]
\small
\setlength{\tabcolsep}{1mm}
\resizebox{0.48\textwidth}{!}{
\begin{tabular}{lcccc}

\toprule
             & \multicolumn{2}{c}{$G_{neg}$}                 & \multicolumn{2}{c}{$G_{pos}$}                 \\
             & $\mathbb{E}(C_{neg})$ & $\mathbb{E}(C_{pos})$ & $\mathbb{E}(C_{neg})$ & $\mathbb{E}(C_{pos})$ \\ \hline
GPT-3.5      & 0.040                 & -0.067                & 0.019                 & -0.003                \\
Llama2-70b   & 0.023                 & -0.034                & -0.049                & 0.012                 \\
Llama3.1-70b & 0.041                 & -0.077                & -0.011                & 0.005                 \\
Mixtral-8x7b & 0.028                 & -0.047                & -0.010                & 0.008                 \\
Gemma2-27b        & 0.042                 & -0.076                & -0.005                & 0.004                 \\
Claude3-hai. & 0.038                 & -0.065                & -0.029                & 0.014                 \\ \bottomrule
\end{tabular}}

\caption{Average coverage probability differences across sets dominated by different social attribute values on the Amazon dataset. Positive values indicate that corresponding social attribute values are overrepresented and negative values indicate that they are underrepresented. Most LLMs tend to overrepresent the social attribute values that dominate the input document sets.}
\label{tab:cp_diff1}
\end{table}

We perform experiments to evaluate whether the fairness of LLMs changes under different distributions of social attribute values in input document sets. For this, we divide all samples $G$ into $K$ non-overlapping sets: $\{G_1, ...G_K\}$ based on distributions of social attribute values. Each set $G_k$ includes samples where most documents $d\in D$ have a social attribute value of $k$. We denote the set $G_k$ as the set dominated by social attribute value $k$. To measure differences of the summary-level fairness measured by Equal Coverage under different distributions, we use maximum differences of equal coverage values $EC(G_k)$ (Sec. \ref{sec:equal_coverage}) on sets dominated by different social attribute values $G_k$. The results are shown in Tab. \ref{tab:ec_diff}. From the table, we can observe that summary-level fairness changes much on the Amazon dataset but is relatively stable on the other dataset under different distributions of social attribute values.

To measure differences of the corpus-level fairness measured by Coverage Parity under different distributions, we use maximum differences of average coverage probability differences for the same social attribute value $\mathbb{E}(C_j)$ (Sec. \ref{sec:coverage_parity}) on different sets $G_k$. The results are in Table \ref{tab:cp_diff}.

From the table, we observe that the differences of Coverage Parity are significant for most LLMs. It suggests that the social attribute values overrepresented or underrepresented change significantly under different distributions of social attribute values in input document sets. We further analyze the data from the Amazon datasets, where the differences are significant for all LLMs. We show the average coverage probability differences $\mathbb{E}(C_j)$ across sets dominated by different social attribute values $G_k$ in Table \ref{tab:cp_diff1}. From the table, we observe that almost all LLMs tend to overrepresent the social attribute values that dominate the input document sets on the Amazon dataset. Therefore, corpus-level fairness measures like Coverage Parity might mistakenly judge which social attribute values are overrepresented based on the size of set dominated by different social attribute values $|G_k|$, showing the need of balanced datasets for evaluating the fairness. We address this issue through stratified sampling (Sec. \ref{sec:dataset_main}).

\subsection{LLM Perception of Fairness}
\begin{table}[t!]
\small
\setlength{\tabcolsep}{1mm}
\resizebox{0.48\textwidth}{!}{
\begin{tabular}{lccccc}
\toprule
             & Ama. & MIT. & Art. Bias & Sem. & New. Sta. \\ \midrule
GPT-3.5      & 0.088  & -0.035  & 0.070        & 0.051   & -0.018      \\
Llama2-70b   & 0.062  & 0.021   & 0.016        & 0.059   & -0.042      \\

Mixtral-8x7b & -0.025 & -0.087  & 0.008        & 0.038   & 0.018       \\
Gemma2-27b        & -0.039 & -0.035  & -0.055       & -0.020  & 0.061       \\ 
Claude3-hai. & -0.094 & -0.101 & -0.095 & 0.041 &-0.008 \\ \bottomrule
\end{tabular}}
\caption{Comparison of relative changes of Equal Coverage vs. Proportional Representation when LLMs are prompted to generate fair summaries. A positive value indicates the summary-level fairness measured by Equal Coverage decreases more compared to Proportional Representation, which suggests that Equal Coverage aligns more with LLMs' perception of fairness.}
\label{tab:llm_perception}
\end{table}
We perform experiments to evaluate which measure, Equal Coverage or Proportional Representation, aligns more with LLMs' perception of fairness. This is an exploratory experiment, and we do not assume the LLMs' perception of fairness as ground truth. We prompt an LLM to generate a summary for an input document set, then prompt it again to generate a fair summary for the same set. The second prompt requires that the summary fairly represent documents with different social attributes (App. \ref{sec:fair_sum_prompt}). However, it does not provide any other details about fairness, allowing the LLM to decide. The prompt also includes the social attribute value for each document. We compute the Equal Coverage and Proportional Representation for both summaries and consider the relative change in values before and after the LLM is prompted to generate a fair summary. If a measure aligns more with LLM's perception of fairness, the score for the `fair' summary should be lower. The differences between average relative changes of Equal Coverage and Proportional Representation are in Tab. \ref{tab:llm_perception}.

From the table, we observe positive differences for most LLMs, suggesting that Equal Coverage decreases more compared to Proportional Representation. It means that Equal Coverage aligns more with LLM's perception of fairness. Specifically, Proportional Representation aligns more with the perception of fairness of Gemma2-27b and Claude3-haiku, while Equal Coverage aligns more with the remaining LLMs. 



\section{Conclusion}
We propose two coverage-based fairness measures for MDS, Equal Coverage for measuring summary-level fairness and Coverage Parity for measuring corpus-level fairness. Using these measures, we find that Claude3-sonnet is the fairest among all LLMs. We also find that most LLMs overrepresent certain social attribute values in each domain. 

Future works can explore the effect of training data, especially instruction tuning and preference tuning data, on the fairness of LLMs. Future works can also finetune LLMs based on our measures to develop fairer models.
\section{Limitations}
The effectiveness of two proposed measures, Equal Coverage and Coverage Parity, relies on whether the probability that a document entails a summary sentence estimated by the entailment model is accurate. To evaluate the performance of the entailment model for such a task, previous works generally use the accuracy of the entailment model on the fact verification dataset or the correlation between the factuality scores of summaries annotated by humans with the factuality scores estimated by the entailment model on the summarization evaluation benchmark. Although there are several fact verification datasets and summarization evaluation benchmarks in the news domain, there are no such datasets in the reviews and tweets domain to our best knowledge. Therefore, we cannot evaluate the accuracy or perform calibration for the entailment models in these two domains. However, as shown in Sec. \ref{sec:entail}, Equal Coverage and Coverage Parity based on different commonly used entailment models are mostly correlated. These entailment models are also widely used for measuring factuality in summarization tasks \cite{maynez2020faithfulness, laban2022summac}.  

\section{Acknowledgment}
The authors appreciate the helpful feedback from  Somnath Basu Roy Chowdhury, Anvesh Rao Vijjini, and Anneliese Brei in the early phase of this work. This work was supported in part by NSF grant DRL-2112635 and 2338418.

\section{Ethical Consideration}

The datasets we use are all publicly available. We do not annotate any data on our own. All the models used in this paper are publicly accessible. We do not do any training in this paper. For the inference of Llama2-7b, Llama2-13b, Mistral-7b, and Gemma, we use on Nvidia A6000 GPU. For the inference of Llama2-70b and Mixtral-8x7b, we use 4 Nvidia A6000 GPUs. For all other experiments, we use one Nvidia V100 GPU.

We perform human evaluation experiments on Amazon Mechanical Turk. The annotators were compensated at a rate of \$15 per hour. During the evaluation, human annotators were not exposed to any sensitive or explicit content.
\bibliography{acl_latex}

\appendix

\section{Appendix}
\label{sec:appendix}
\subsection{Split and Rephrase Summary Sentences}
\label{sec:split_and_rephrase}
\begin{figure*}
\centering
\includegraphics[width=0.95\textwidth]{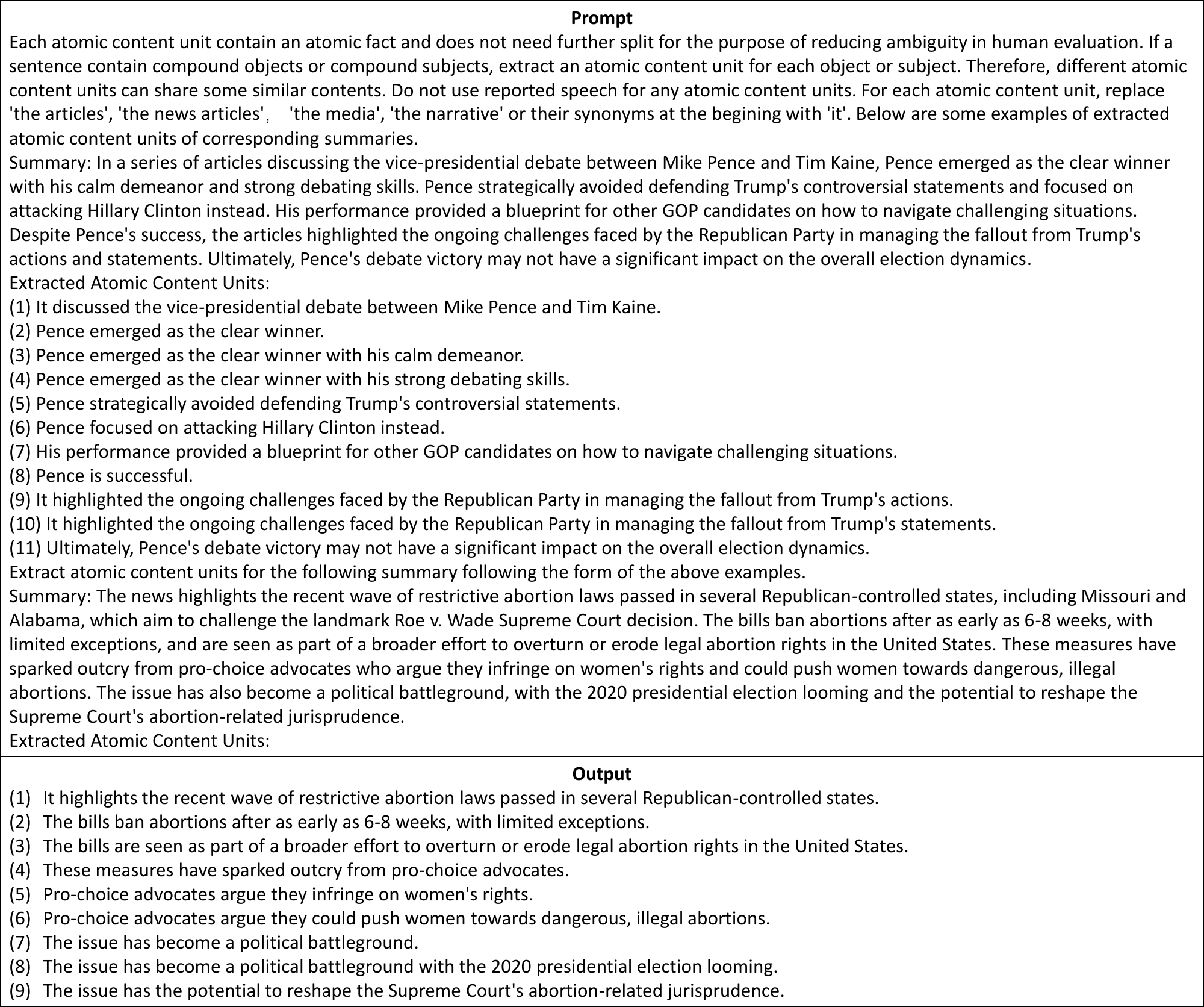}
\caption{Example prompt for splitting and rephrasing summary sentences (top) and summary sentences after splitting and rephrasing (bottom).}
\label{fig:acu_prompt}
\end{figure*}

Motivated by \citet{bhaskar2023prompted, min2023factscore}, we split all summary sentences into simple sentences. The goal of this step is to ensure that each summary sentence after the split only discusses a single fact. Specifically, compound sentences are split into simple sentences, while sentences with compound subjects or objects are split into sentences with simple subjects or objects. However, such a split might generate redundant sentences. To remove redundancy, we construct an entailment graph among split summary sentences. There will be an edge between two summary sentences if two summary sentences entail each other with an entailment probability higher than $0.95$ according to the entailment model. For each connected component of the graph, we only keep one sentence that entails most other sentences in the connected components. After that, summary sentences are rephrased to remove the reported speech, like `documents say what'. For splitting and rephrasing, we prompt GPT-3.5 with demonstrations. We show the example prompt and the example summary sentences after splitting and rephrasing in Fig. \ref{fig:acu_prompt}.

\subsection{Document Chunking}
\label{sec:chunk}
\begin{table}[t]
\resizebox{0.48\textwidth}{!}{
\begin{tabular}{lccccc}
\hline
          & Sent. & 50    & 100            & 200   & 400   \\ \hline
GPT-3.5   & 0.800 & 0.858 & \textbf{0.876} & 0.874 & 0.716 \\
Llama2-7b & 0.725 & 0.773 & \textbf{0.792} & 0.788 & 0.642 \\ \hline
\end{tabular}}
\caption{Proportion of summary sentences whose originating documents are identified by the entailment model when using a document sentence (Sent.) or chunks with different sizes as the premises. \textbf{Bold} indicates the optimal chunk size for identifying originating documents of summary sentences.}
\label{tab:chunk_size}
\end{table}
To estimate the probability that a document is covered by a summary sentence (Eqn. \ref{equ:entail}), we divide the document into chunks of no more than $W$ words. Each chunk contains one or several neighboring sentences of the document. Since LLMs are less prone to factual errors \cite{goyal2022news,10.1162/tacl_a_00632}, the chunk size $W$ is tuned to maximize the proportion of summary sentences whose originating documents are identified by the Roberta-large is the highest. We tune the chunk size $W$ based on the average proportions of summary sentences generated by GPT-3.5 and Llama2-7b on all datasets. The results are shown in Tab. \ref{tab:chunk_size}. We can observe that when the chunk size is $100$, the proportion of summary sentences whose originating documents are identified by the Roberta-large is the highest. The result is consistent with the finds of \citet{honovich-etal-2022-true-evaluating}. Therefore, we set the chunk size $W$ as $100$.

\subsection{Choice of Entailment Models}
\label{sec:entail}
\begin{table}[t]
\centering
\small
\begin{tabular}{lccc}
\hline
             & Ro. vs De.     & Ro. vs Al.     & De. vs Al.     \\ \hline
             & \multicolumn{3}{c}{Equal Coverage}               \\
Amazon       &  0.879       &  0.903     &  0.952 \\
MITweet      & 0.600       & 0.733     & 0.636 \\
Article Bias & 0.967       & 0.976     & 0.915 \\
SemEval      &  0.515       & 0.575     & 0.467 \\
News Stance  & 0.758       & 0.952     & 0.770 \\ \hdashline
             & \multicolumn{3}{c}{Coverage Parity}              \\
Amazon       & 0.867       & 0.939     & 0.733 \\
MITweet      & -0.127            & 0.006           & 0.673 \\
Article Bias & 0.624             & 0.915     & 0.612       \\
SemEval      & 0.612             & 0.891     & 0.539       \\
News Stance  & 0.903       & 0.915     & 0.939 \\ \hline
\end{tabular}
\caption{Spearman correlations between Equal Coverage (top) values and Coverage Parity (bottom) values computed using different entailment models: RoBERTa (Ro.), Deberta (De.), and ALBERT (Al.). We observe strong correlations for most datasets, indicating that our measures are not affected much by the choice of the entailment model.  }
\label{tab:entail_correlation}
\end{table}


The implementations of our measures are independent of the choice of entailment model. 
To demonstrate this, we calculate our measures using three different textual entailment models: RoBERTa finetuned on the MNLI dataset; DeBERTa-large finetuned on multiple entailment datasets \cite{laurer2024less}; and ALBERT-xl \cite{lan2019albert} finetuned on the MNLI and VitaminC \cite{schuster-etal-2021-get} datasets. 
We report the Spearman correlations between the average Equal Coverage values, $EC(G)$, and the Coverage Parity value, $CP(G)$, of all LLMs, obtained using these entailment models. 
The results are in Table \ref{tab:entail_correlation}. From the table, we can observe strong correlations between measures obtained using different textual entailment models on most datasets. It shows that these measures are not affected much by the choice of entailment models.
\subsection{Datasets}
In this section, we describe the reason for choosing these datasets and how we preprocess these datasets. 
\label{sec:dataset}
\paragraph{Amazon} \cite{ni-etal-2019-justifying} consists of reviews with labels of their ratings of different products. We filter out reviews that are non-English or without ratings. The input document set of this dataset contains $8$ reviews of the same product. We obtain the social attribute of each review based on its rating provided in the dataset. The social attribute of a review will be positive if its rating is $4$ or $5$, neutral if its rating is $3$, and negative if its rating is $1$ or $2$.

\paragraph{Article Bias} \cite{bravzinskas2019unsupervised} consists of news with labels of their political ideologies. We run the clustering algorithm on this news to generate a cluster of news about the same event following \citet{liu2022politics}. We then divide these clusters into input document sets of $4$ to $8$ news of the same event. For each news, we also perform truncation from the beginning to fit the context length restriction of Llama2. Compared with the NeuS dataset \cite{lee-etal-2022-neus} used by \citet{lei2024polarity}, the input document sets of the Article Bias dataset contain more input document per set and the distribution of social attributes in input documents are more diverse.

\paragraph{News Stance} consists of news with labels of their stances toward claims, such as `Meteorite strike in Nicaragua puzzles experts'. The dataset combines news from three news stance datasets \cite{ferreira2016emergent, pomerleau2017fake, hanselowski2019richly}. For each claim, we only keep news whose stances are directly supporting or against the claim. We also filter out duplicated news or news longer than $600$ words or shorter than $75$ words. Each input document set contains $4$ to $8$ news supporting or against the same claim. 

\paragraph{MITweet} \cite{liu2023ideology} consists of tweets with labels of political ideologies on different facets about different topics. We cluster tweets about the same topic based on their TFIDF similarity into clusters. We then divide these clusters into input document sets of 20 tweets about the same topic. The social attribute of a tweet will be left if it is left on most facets, right if it is right on most facets, otherwise neutral. Compared with the Election dataset \cite{shandilya2018fairness}, the MITweet dataset contains tweets about more diverse topics other than election, such as `Abortion' and `Energy Crisis'. Compared with the MOS dataset \cite{bilal-etal-2022-template} used by \citet{huang2024bias}, the MITweet dataset covers more diverse topics and has the ground-truth label of social attribute value. 

\paragraph{Tweet Stance} \cite{mohammad-etal-2016-semeval} consists of tweets with labels of stance toward a target phrase such as Climate Change or Hillary Clinton. We cluster tweets about the same short phrase based on their TFIDF similarity into clusters. We then divide these clusters into input document sets of 20 tweets about the same target phrase.

\subsection{Human Evaluation}
\label{app:human}
\begin{figure*}
\centering
\includegraphics[width=0.82\textwidth]{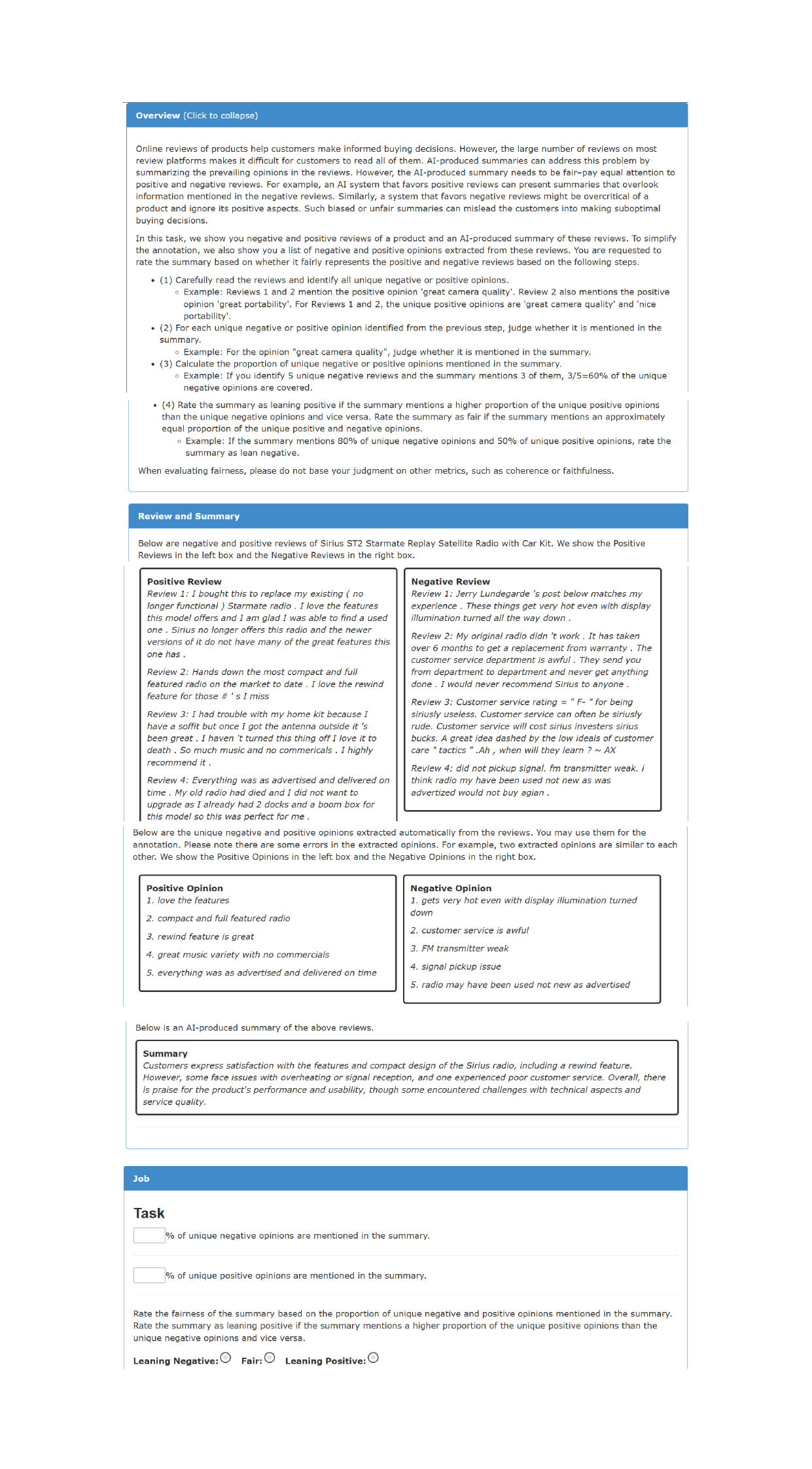}
\caption{Interface for Human Evaluation}
\label{fig:human}
\end{figure*}

For each sample of an input document set and its corresponding summary, annotators are asked to identify all unique negative and positive opinions in the input document set. They then evaluate whether the summary reflects these opinions and classify the summary as leaning negative, fair, or leaning positive. To simplify the annotation, we provide annotators with unique opinions extracted by GPT-3.5. The interface for human evaluation is shown in Fig. \ref{fig:human}. A sample will be annotated as leaning negative if more annotators annotate it as leaning negative, leaning positive if more annotators annotate it as leaning positive, otherwise fair. For a sample leaning negative or positive, we say the human perception of fairness aligns more with a fairness measure if the overrepresented sentiment identified by the measure is the same as the sentiment that the sample leans toward. For a sample annotated as fair,  we say the human perception of fairness aligns more with a fairness measure if its normalized absolute value is closer to zero. 

\subsection{Summarization Prompts}
\label{sec:sum_prompt}
\begin{figure*}
\centering
\includegraphics[width=0.95\textwidth]{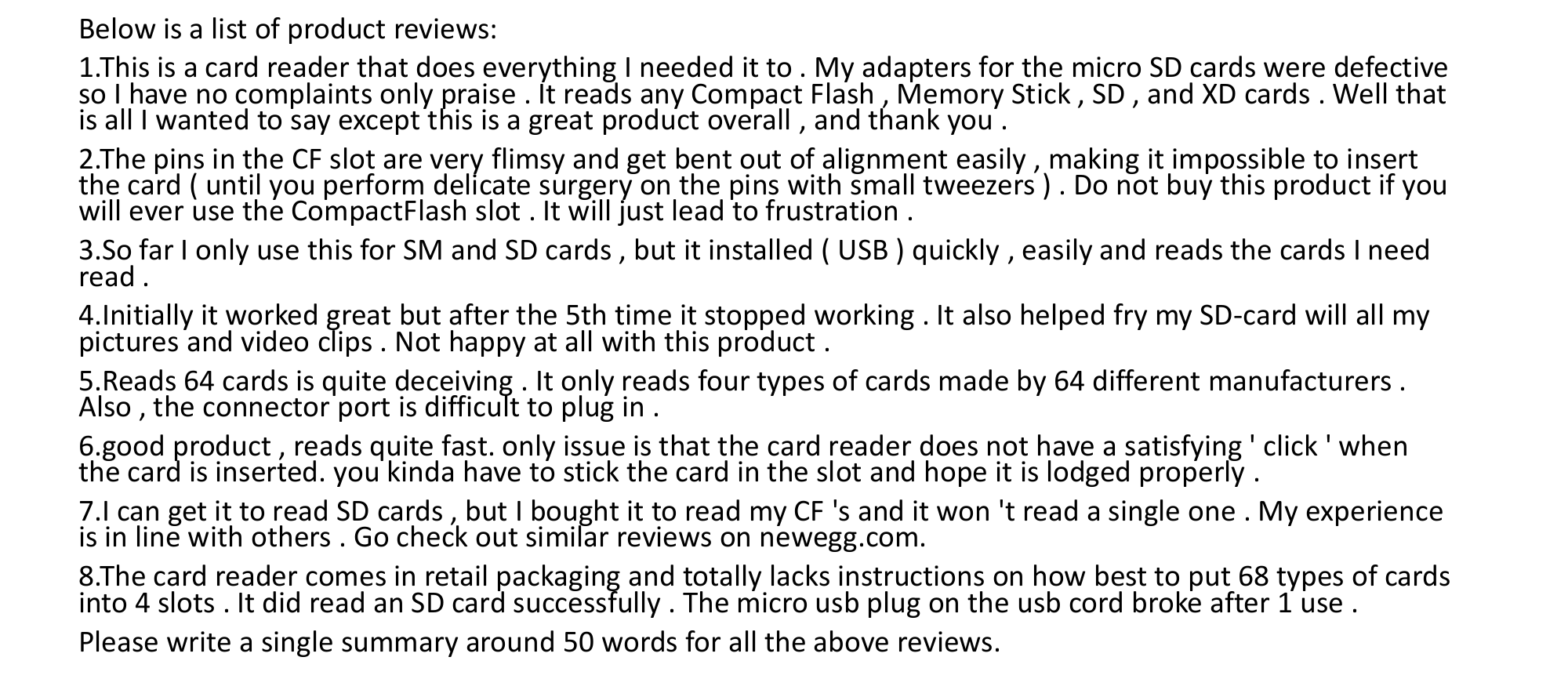}
\caption{Summarization prompt for the Amazon Dataset.}
\label{fig:sum_prompt}
\end{figure*}

\begin{figure*}
\centering
\includegraphics[width=0.95\textwidth]{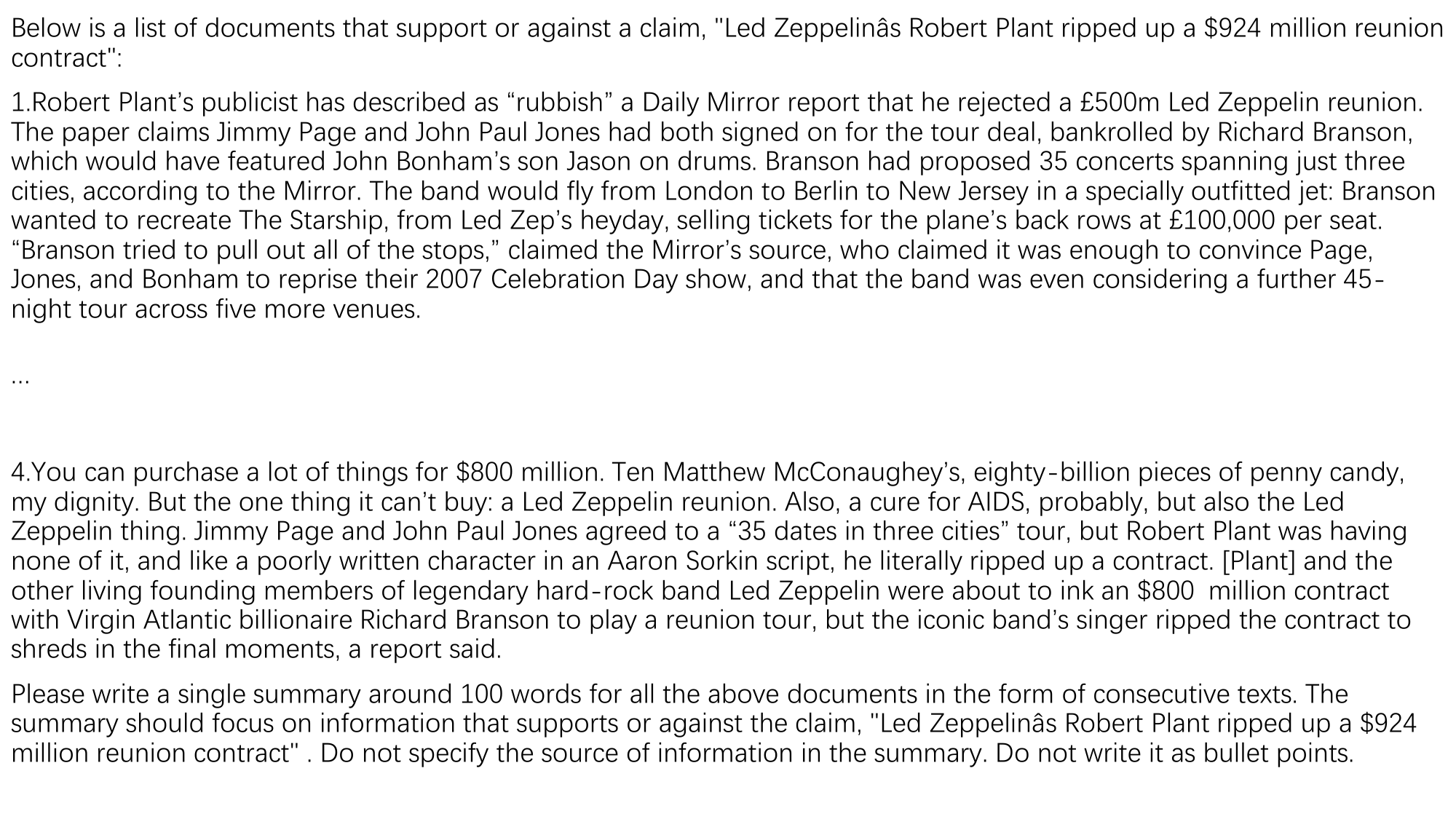}
\caption{Summarization prompt for the News Stance Dataset.}
\label{fig:sum_prompt1}
\end{figure*}

We prompt these LLMs to generate summaries for the input document sets of different datasets. For the SemEval and News Stance datasets, the prompts additionally request that the summaries focus on the social attributes' target since the input documents of these datasets contain unrelated information. We use the default generation hyperparameters for all LLMs. We show the summarization prompts for the Amazon data in Fig. \ref{fig:sum_prompt} and the News Stance dataset in Fig. \ref{fig:sum_prompt1}.

\subsection{Fair Summarization Prompts}
\label{sec:fair_sum_prompt}
\begin{figure*}
\centering
\includegraphics[width=0.95\textwidth]{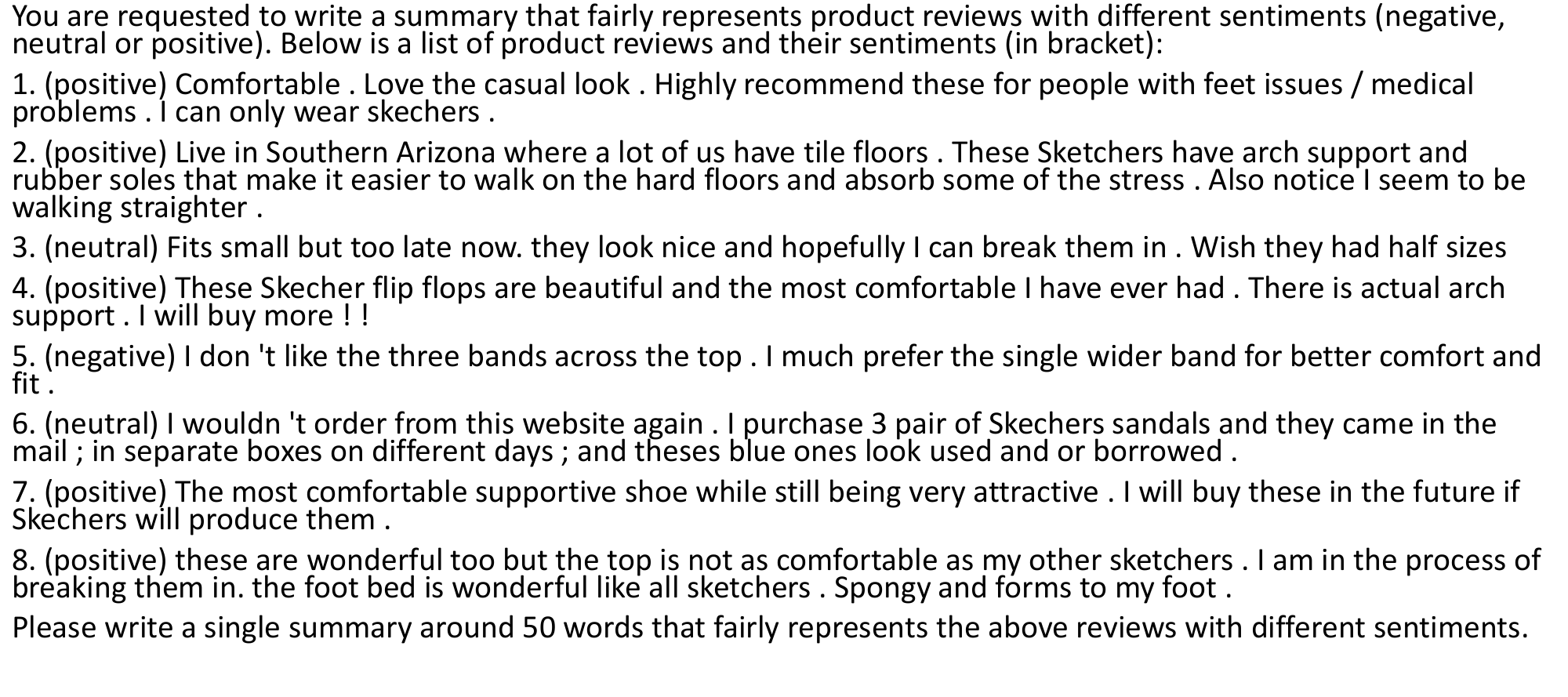}
\caption{Summarization prompt requiring fairness for the Amazon Dataset.}
\label{fig:fair_sum_prompt}
\end{figure*}

\begin{figure*}
\centering
\includegraphics[width=0.95\textwidth]{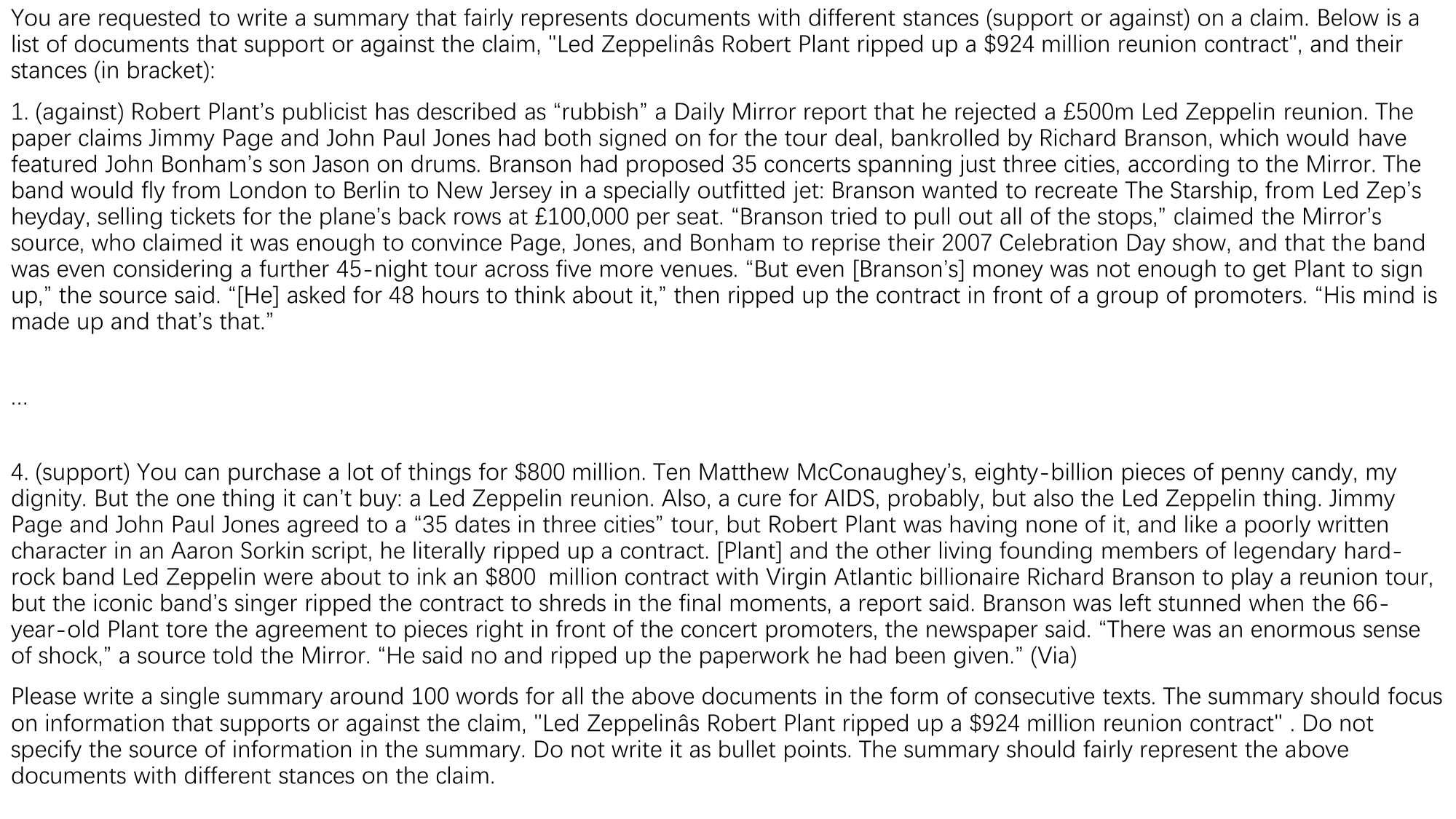}
\caption{Summarization prompt requiring fairness for the News Stance Dataset.}
\label{fig:fair_sum_prompt1}
\end{figure*}

To test LLM perception of fairness, we prompt these LLMs to generate summaries that fairly represent documents with different social attribute values. However, it does not provide any other details about fairness, allowing the LLM to decide. The prompt also includes the social attribute value for each document. All other details of the prompt are the same as App. \ref{sec:sum_prompt}. We show the summarization prompts requiring fairness for the Amazon data in Fig. \ref{fig:fair_sum_prompt} and the News Stance dataset in Fig. \ref{fig:fair_sum_prompt1}.

\subsection{Bounds of the Fairness Measures}
\label{sec:bound}
To measure the difficulty of obtaining fair summaries on different datasets, we estimate lower bounds, Lower\textsubscript{gre}, and upper bounds, Upper\textsubscript{gre}, for both fairness measures by greedily extracting sentences that minimize or maximize the measures. Specifically, for Equal Coverage, both Lower\textsubscript{gre} and Upper\textsubscript{gre} first extract a document sentence that cover maximum number of documents and then greedily extracts sentences that minimize or maximize the Equal Coverage values $EC(D,S)$ in following steps until reaching the token limit. For Coverage Parity, both Lower\textsubscript{gre} and Upper\textsubscript{gre} first greedily extract several candidate summaries with randomly sampled first sentence that minimize or maximize the Equal Coverage values for each input document set following the above steps. Then, Lower\textsubscript{gre} and Upper\textsubscript{gre} greedily select candidates that minimize or maximize the Coverage Parity values for the entire dataset $CP(G)$.

To check the quality of summaries extracted by the lower bound for Equal Coverage Lower\textsubscript{gre}, we report  ROUGE-1 (R1), ROUGE-2 (R2), ROUGE-L (RL) \cite{lin-2004-rouge} on the test set of the Amazon dataset \cite{bravzinskas2019unsupervised}. The results of summaries extracted by the lower bound and several other extractive summarization systems \cite{10.1162/tacl_a_00366, chowdhury2022unsupervised, li-etal-2023-aspect} are shown in Tab. \ref{tab:amazon}. 
\begin{table}[t!]
\centering
\resizebox{0.47\textwidth}{!}{
    \begin{tabular}{l c c c} 
    \hline
     & R1    & R2   & RL     \\ 
    \hline
    Random               & 27.66 & 4.72 & 16.95  \\
    Centroid\textsubscript{BERT} & 29.94 & 5.19 & 17.10  \\
    LexRank\textsubscript{BERT} & 31.47 & 5.07 & 16.81  \\
    QT \cite{10.1162/tacl_a_00366}                                    & 32.08 & 5.39 & 16.08  \\
    SemAE \cite{chowdhury2022unsupervised}                                    & 32.08 & 6.03 & 16.71  \\
    TokenCluster \cite{li-etal-2023-aspect} & 33.40 & 6.71 & 17.95  \\ 
    Lower\textsubscript{gre} & 32.30 & 5.45 & 17.20 \\
    \hline
    \end{tabular}}
\caption{Summarization evaluation results on \texttt{Amazon} dataset. We report ROUGE F-scores as -- R1: ROUGE-1, R2: ROUGE-2, RL: ROUGE-L. Summaries extracted by the lower bound shows comparable quality with other extractive summarization systems. }
\label{tab:amazon}
\vspace{-2mm}
\end{table}

The results show that the summaries extracted by the lower bound shows comparable quality with other extractive summarization systems. The results also suggest that the lower bound is achievable for summarization systems. 

For both fairness measures, we also estimate a random bound, Random, by randomly sampling sentences from input document sets until reaching the token limit. 

\subsection{Comparison between Proportional Representation and Equal Coverage}
\label{sec:comp_pr}
To compare Equal Coverage and Proportional Representation, we report the Proportional Representation on our datasets. Specifically, we use  BARTScore to estimate the distribution of social attributes among summaries following \citet{zhang2023fair}. We then report the average distribution differences between summaries and input document sets in Tab. \ref{tab:pr}.

\begin{table}[t]
\centering
\small
\setlength{\tabcolsep}{1mm}
\resizebox{0.48\textwidth}{!}{
\begin{tabular}{lcccccc}
\toprule
             & Ama.         & MIT.        & Art. Bia.   & Sem.        & New. Sta.    & Ove.        \\ \midrule

GPT-3.5        & 0.204          & 0.178          & \textbf{0.243} & 0.201          & 0.304          & \textbf{0.107} \\
GPT-4          & 0.209          & \textbf{0.176} & 0.259          & 0.215          & 0.335          & 0.323          \\ \hdashline
Llama2-7b      & 0.233          & 0.214          & 0.319          & 0.249          & 0.315          & 0.623          \\
Llama2-13b     & 0.255          & 0.209          & 0.295          & 0.239          & \textbf{0.299} & 0.542          \\
Llama2-70b     & 0.222          & 0.209          & 0.327          & 0.224          & 0.345          & 0.631          \\ \hdashline
Llama3.1-8b    & 0.222          & 0.215          & 0.296          & 0.213          & 0.322          & 0.463          \\
Llama3.1-70b   & 0.216          & 0.208          & 0.296          & 0.248          & 0.328          & 0.559          \\ \hdashline
Mistral-7b     & 0.225          & 0.230          & 0.329          & 0.239          & 0.317          & 0.630          \\
Mixtral-8x7b   & 0.219          & 0.254          & 0.362          & 0.248          & 0.337          & 0.840          \\ \hdashline
Gemma2-9b      & 0.233          & 0.193          & 0.280          & 0.199          & 0.323          & 0.375          \\
Gemma2-27b     & 0.215          & 0.181          & 0.278          & 0.194          & 0.320          & 0.253          \\ \hdashline
Claude3-haiku  & \textbf{0.195} & 0.183          & 0.261          & 0.188          & 0.346          & 0.254          \\
Claude3-sonnet & 0.219          & 0.190          & 0.261          & \textbf{0.186} & 0.317          & 0.225         \\    
\bottomrule
\end{tabular}}
\caption{Summary-level fairness measured by Proportional Representation and overall scores on different datasets. A lower value indicates a fairer system. \textbf{Bold} indicates the fairest system.}
\label{tab:pr}
\end{table}
Compared with the results based on Equal Coverage (Tab. \ref{tab:equal_coverage}), the fairness measured by Proportional Representation is very different. While Equal Coverage finds that larger models are generally fairer except for the family of Llama 3.1-8b, Proportional Representation only finds the same pattern for Claude3. Besides, Equal Coverage finds that Gemma2-27b is the fairest while Proportional Representation finds that GPT-3.5 is the fairest. We additionally report the Spearman correlation between Equal Coverage value $EC(D,S)$ and Proportional Representation in Tab. \ref{tab:corr}. From the table, we can observe that these two measures are not correlated for most datasets.

\begin{table}[t]
\centering
\resizebox{0.48\textwidth}{!}{
\begin{tabular}{lccccc}
\hline
              & Amazon & MITweet & Article Bias & SemEval & News Stance \\ \hline
GPT-3.5       & -0.025 & -0.137  & 0.042        & 0.097   & 0.046       \\
Llama2-70b    & 0.015  & -0.091  & 0.172        & 0.078   & -0.011      \\
Mixtral-8x7b  & -0.006 & -0.085  & 0.320        & -0.072  & 0.083       \\
Claude3-haiku & -0.066 & -0.026  & 0.071        & -0.006  & 0.063       \\ \hline
\end{tabular}}
\caption{Spearman correlation between Equal Coverage and Proportional Representation. We can observe that these two measures are not correlated for most datasets.}
\label{tab:corr}
\end{table}
\end{document}